# Similarity among the 2D-shapes and the analysis of dissimilarity scores

**Karel ZIMMERMANN**
28170 Neuville-la-Mare, France
*kzimm007@gmail.com*

**Abstract**
We present a conceptually simple and intuitive method to calculate and to measure the dissimilarities among 2D shapes.
Several methods to interpret and to visualize the resulting dissimilarity matrix are presented and compared.

## 1. INTRODUCTION

### 1.1 Shapes similarity

Who hasn't seen the jigsaw puzzle the retired persons assembly on gloomy afternoons, or a baby playing with the shape sorting cube? And everybody remembers from school that Italy is a "boot" and the young French learn that France is a "Hexagone". But what means the shape similarity of two figures, how to calculate and how to measure that?

Some approaches (e.g., in biology) use various shape descriptors (circularity, roundness, convexity etc.). They consider two shapes similar if they have similar descriptors. However, two pieces of a jigsaw puzzle might have even identical descriptors, but very different shapes. The methods of the face recognition [1] use the fact that each human face can be described by the same number of "landmark" points (eyes, mouth, nose...). But how many points to use and how assign them each to other between two quite general shapes? Some authors try to solve even such a problem [2]. However, it is not quite clear if that approach could be used for any shapes, e.g., non-contiguous (with "islands") or with "holes" inside.

In a preceding paper [3] we have used a quite different and intuitive approach: imagine two figures A and B cut of a cardboard and put one over another. They will surely mutually overlap and there will be some non-overlapping parts. In the computer the 2D shapes as well as the non-overlapping parts are represented by polygons and one can calculate their surfaces. One can than move the "mobile" figure B and modify its size to minimize the total surface of all the non-overlapping parts:

$$s(A, B) = \min(\Delta A + \Delta B) \qquad (1)$$

where $\Delta A$ is the total surface of all the non-overlapping parts of the figure, and similarly for $\Delta B$. In reality, s(A,B) measures the dissimilarity: it is zero when the figures A and B are identical and the figure B is attentively posed on A, otherwise it is positive. In this way we have found that the French "Hexagon" resembles better to the square than to the hexagon.

The concept of similarity is closely related to the operations we use. In general, any shape deformation and movement could be used provided that it could be mathematically well described. In the jigsaw puzzle it is translation and rotation only, neither the size variation, nor the flip. When comparing the car shapes, no one would probably use the rotation to compare two cars oriented in orthogonal or opposite directions or a car with the other upside down. In the "France-hexagon" problem [3], we have used the translation, rotation as well as the size variation. In this paper we use the same set of geometrical transformations.

We present here a new version of the formula (1), adapted to measure the dissimilarity among a set of shapes.

### 1.2 Analysis of the dissimilarity score matrix

If one calculates the pairwise scores (dissimilarity, distance etc.) among a set of objects, it is interesting to use it to obtain an information about the inner structure of that set: to class the objects, to map them and to visualize it.

The seemingly simplest is to cluster the close objects using only the obtained scores. The amino acid substitution score matrices [4], with their positive and negative entries and non-zero diagonals, are far from being the distance matrix and having any evident structure. However, by permuting manually lines and corresponding columns, J-F. Gibrat obtained an approximative block diagonal matrix. The obtained diagonal blocs corresponded well to the usual amino acid similarity classification maps [5]. Later he has developed this idea using the matrix algebra [6]. We present here the approach which imitates the original manual procedure.

If the scores can be assimilated to distances, various methods exist to obtain cartesian coordinates and to map the objects. They are known under the name MDS (Multidimensional Scaling) [7] . The "royal way" seems to use the Torgerson matrix [8] and then to use K-means method [9]. The seeming advantage of the Torgerson matrix is that it avoids any function minimization. However, we don't suffer of the "horror minimalizationis" and by using the numerical minimization we obtain better results.

A remark should be made concerning the score matrices which cannot be assimilated to distances, as e.g., the case of the amino acid substitution matrices mentioned above. One possibility for such "problematic" matrices is to consider the matrix entries not as distances but as the scalar products of euclidean vectors [10]. But, when the diagonal is "sacrificed", another possibility is to search the euclidean vectors such that the entries of the distance matrix *d* correlate (or anti-correlate, depending of the nature of the data) with the corresponding entries of the data matrix *D* [10]). We present this approach here, too, and compare it to the other methods.

## 2. METHOD

### 2.1 Dissimilarity score

The formula (1) is good for fitting various ("mobile") shapes B to the same "fixed" shape A (we keep always the first of the two shapes in the formula $s(A, B)$ fixed and fit the second to the first one in order to prevent any drift in the minimization and to reduce the number of parameters). But it makes that the formula (1) is not symmetric and thus it cannot be used to calculate pairwise similarities among a set of objects: fitting a small B to the big A doesn't yield the same result as fitting the big A to the small B. Noting $\Delta A, \Delta B$ as above the surfaces of the non-overlapping parts, we have thus defined the dissimilarity score :

$$s(A, B) = \min_{parB} 100 \frac{\Delta A + \Delta B}{A + B} \qquad (2)$$

Here *A, B* represent the surfaces of the corresponding objects and *parB* the set of the parameters of the geometrical transformation of B: the size, the position (x,y) and the orientation (angle α). The dissimilarity score s(A,B) is strictly non-negative, between 0% and 100%. The formula (1) is well symmetric within the precision of the minimizer used. It might seem that increasing the size of B might reduce ΔA and thus the value of *s(*A,B*)*. But, in such a case, ΔB prevents B to increase too much. On the other hand, decreasing B increases ΔA. In this way there is always an optimal size of B and the results are very close to that of the formula (1). One might think about some other metric, using, e.g., the squares of the surfaces ΔA and ΔB. That would reduce big non-overlapping parts at the price of the small ones. But it might, too, make the function s(A,B) more complicated and thus hinder the convergence of the minimization. We have not used that metric in this work. The only constraint on the variables is the positivity of the size parameter, which is easy to respect. That is why nearly any ready-made nonlinear minimizer can be used to compute s(A,B). It is evident that the gradient of the objective function s(A,B) can be computed only numerically. For our 4 parameters of the minimization the gradient represents 8 computations of the objective function s(A,B), the Hessian would represent 48 computations. That is why we have written in Scilab [11] a rudimentary truncated quasi-Newton limited memory [12] minimizer.

A really serious problem are many local minima. It is thus necessary to repeat the minimizations with various initial orientations α of the mobile figure B. A lengthy work was to obtain "manually" the (x,y) coordinates of the contours of the figures. Several photo editors (Photoshop, Gimp, Zoner Photo Studio) can be used for that.

### 2.2 Analysis of the results

The methods presented and compared below are quite general, they can be used for the dissimilarity data as well as for the distances. That is why we'll speak not only of figures or shapes but more generally of objects, and of distances or scores data. We'll note *D* the $N \times N$ the score (data) matrix (with the entries $D_{i,j}$ and *d* its calculated approximation (with the entries $d_{i,j}$).

To illustrate and to compare the methods, we use two examples: a "known" one (road distances) and then the dissimilarity scores.

#### 2.2.1 Block matrix clustering

The idea is to group together the similar (close) entries of the data (scores) matrix by minimizing the sum of intra-cluster scores. The method tries to imitate the Gibrat's original manual procedure mentioned above. The clusters can be then visualised as the blocks on the matrix diagonal, but it is not necessary.

Let denote *K* the user defined number of clusters and *X* the $K \times N$ the matrix of variables. Its binary 0/1 entries $X_{k,n}$ ($k = 1...K; n = 1...N$) are equal to 1 if the $n$−th object is in the cluster k and $X_{k,n} = 0$ if not. Every object must be included in one and only one cluster, thus $\sum_{k=1}^{K} X_{k,n} = 1$ for all $n = 1...N$

Let define the squared binary *selection matrix* $Q = X^T \cdot X$ (the superscript *T* means transposition) and the $N \times N$ binary *block-matrix B*, $B_{m,n} = Q_{m,n} D_{m,n}$ for all $m = 1...N, n = 1...N$. The problem is then to find the optimal matrix *X* by minimizing the sum of the entries of the block matrix *B*:

$$X_{\text{opt}} = \operatorname*{argmin}_{X} \sum_{m=1}^{N} \sum_{n=1}^{N} B_{m,n} \qquad (3)$$

with the constraint $\sum_{k=1}^{K} X_{k,n} = 1$ for all $n = 1...N$. The minimization problem is a non-linear (the matrix *Q* is a product) integer (binary) programming problem.

In practice, during the minimization, the only 1 in each of the *N* columns of the matrix *X* jumps among the *K* lines to find its optimal position.

For the tests with 35×35 data matrix $D$ and $K \leq 5$ the non-linear minimizer Solver of Excel still worked (it is limited up to 200 variables). In reality it solves a relaxed problem and the resulting variables are not exactly binary and must be rounded at the end of the procedure. That is why we have written in *Scilab* a rudimentary genetic algorithm. Because of the local minima it is necessary to restart the problem with different random initial settings of the matrix $X$, but otherwise the program does well the job.

One might also think to constitute the clusters by minimizing the squares of the intra-cluster distances. Technically it doesn't make any difficulty but it would represent a different problem.

It might be interesting to determine a *centroid* of each cluster. As the program doesn't use cartesian coordinates, only the existing objects can be used as the centroids. It seems straightforward to choose for the centroid the element which minimizes the sum of distances between the centroid and all other elements in the same cluster. However, this would correspond to the "median" centroid. That is why we use the more habitual "mean" centroid which minimizes the sum of the squares of the distances. If there is only one element in the cluster, it represents automatically the centroid, if there are only two elements, any of them can be considered as the centroid of the cluster. On the contrary, with the increasing number of elements in the cluster, the centroid approaches its mass center.

### 2.2.2 GMDS-Generalized Multidimensional Scaling

All is easier if one can have the cartesian coordinates of the objects. A direct approach, using minimization, is called GMDS (GMDS-Generalized Multidimensional Scaling) [7] : the cartesian $n$-dimensional position vectors $\vec{x_i} = (x_{i,1}, x_{i,2}, \dots x_{i,n})$ of the objects can be obtained by minimizing the difference between the calculated distances $d$ and the corresponding data matrix $D$:

$$\underset{\vec{x_1},\dots \vec{x_N}}{\mathrm{argmin}} \sum_{i=1}^{N} \sum_{j>i}^{N} (D_{i,j} - d_{i,j})^2 \quad (4)$$

where $d_{i,j} = \|\vec{x_j} - \vec{x_i}\|$ .

The dissimilarity scores are between 0% and 100%, it seems that the distances should be constrained by 100, as gladiators in a circular arena with diameter 100. However, it is hard to imagine two figures which cannot overlap at all, so the formula (2) yields in practice always the scores much smaller than 100. In our examples we have used the Excel Solver non-linear minimizer. To prevent the minimizer to drift we have used constraints $\sum_{m=1}^{N} X_{m,i} = 0, \ i = 1 \dots n$ , but it is not essential.

### 2.2.3 Torgerson matrix method

This method to obtain the cartesian coordinates is well described elsewhere [8]. It uses SVD (Singular Value Decomposition) to obtain the cartesian coordinates $\vec{x_i}$ which reconstitute best the Torgerson matrix and, as a "by-product", the score matrix $D$, too. It is theoretically elegant and avoids the numerical minimization. However, it should be noted that the Torgerson matrix is a nonlinear transformation of the original data matrix. The method thus solves a different problem. Another problem is the dimensionality of the resulting position vectors. The corresponding vectors U and V of the SVD of a symmetric matrix are necessarily either identical or they have opposite signs. With real matrices there are nearly always several terms with opposite signs. These terms can still be used to reproduce the Torgerson matrix, but cannot be used to calculate the cartesian coordinates. Thus, with real data matrices, the cartesian coordinates are calculated with a limited number of terms of the SVD. To make the SVD of the Torgerson matrix, we have used the ready-made program in Scilab or we have made it by minimization using the Excel Solver.

### 2.2.4 K-means

This known clustering method is well described elsewhere [9]. Its difference with the other methods is discussed below. We have written this method in a simple program in Scilab.

### 2.2.5 Correlation method

The main significance of this method is in the situations when the data $D$ cannot be easily assimilated to distances (as, e.g., the amino acid substitution matrices mentioned above with negative entries) and the calculated euclidean distances $d$ cannot be directly compared to the data matrix $D$. In such situations one can try to find the cartesian coordinates of the objects such that the calculated distance matrix $d$ correlate best with the data matrix $D$:

$$\max_{parB} r(D; d) \quad (5)$$

where $r$ is the correlation coefficient and *parB* means the parameters (size, position, orientation) of the figure B. In our examples we have used the Excel Solver minimizer.

### 2.2.6 Graphical output

Most figures were made with Excel and PowerPoint programs. The 3D scatter graphs were made with Mathematica.

## 3. RESULTS AND DISCUSSION

### 3.1 Data

It is always easier to present any method with smartly chosen simulated data. However, we have decided to use a less easy way. To verify the method we have used the European road distances matrix among 35 European towns [13], for which the results can be compared with the real situation. The data should represent essentially distances in 2D space, but we have found that in 510 cases (from 595 distances) they violate the triangle inequality. The matrix entries spread continuously over the whole interval between the minimal (120 km) and the maximal distance (3902 km), the matrix does not seem to have any particular structure.

For the shape dissimilarity, we have used 25 states (Fig. 1) that we have come across on internet.

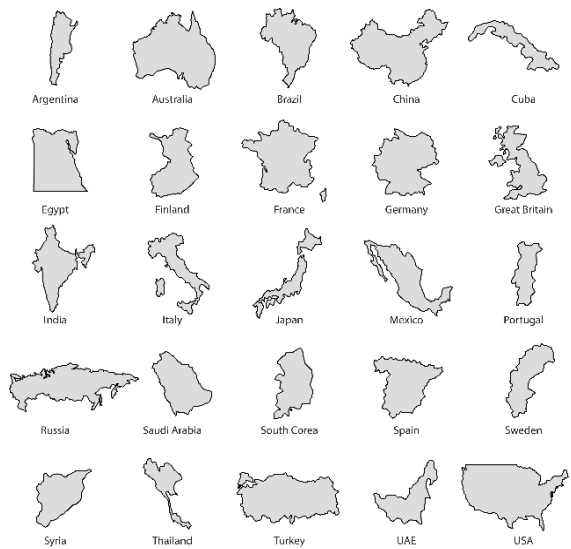

Fig.1 The states used as 2D shapes. The states are not to scale.

We precise here that, even if we use the country names, these figures are only inspired by the existing countries and in any case, they do not pretend to represent the real country borders and even less the geopolitical opinion of the authors.

The mean number of points noted on the contours was $136 \pm 52$. The maximum was for Russia (290 points), the minimum for UAE and Syria (68 points). Many figures are non-convex and non-contiguous. None of them contain "holes" inside (as is the case, e.g., of the South Africa, not included), but it wouldn't represent any problem. We were surprised that there are no violations of the triangle inequality among the calculated dissimilarity scores and thus they can well be assimilated to distances, better than the road distances. The scores spread continuously over the whole interval between the minimum (10.3 %) and the maximum (52.7 %), the data matrix does not seem to have any particular structure. The mean dissimilarity score is $27.1 \pm 5.7$ %. The three most similar shapes are Australia-Germany (10.3 %), Saudi Arabia-USA (10.6 %) and Egypt-Saudi Arabia (10.9 %). The most dissimilar is the pair France-Japan (52.7 %). We have sent the figure 1 to many friends asking them to write us their perception of the similarities. The first reaction of nearly all the people was that couldn't find any similarity. When we insisted, 23 % of people considered as the "most similar" the pairs Portugal-Sweden (calculated dissimilarity score 14.6 %), 10 % Mexico-Cuba (score 33.2 %) and 10 % China-Australia (score 16.9 %). It seems probable that the first two pairs were selected clearly for their elongated shapes. Nobody has indicated any of the three pairs with the smallest dissimilarity score. Either the human perception of similarity is different or the task was too hard without the possibility to try with real figures as in the jigsaw puzzle.

### 3.2 Data analysis

#### 3.2.1 Block matrix clustering

3.2.1.1 Road distances data

For the number of clusters $K = 2...5$ the results are shown in the figure 2.

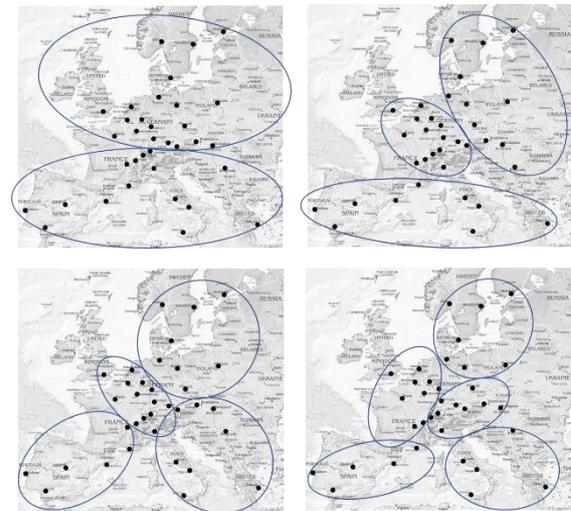

Fig. 2: Block matrix clustering of the road distances matrix for the number of clusters K = 2. . . 5. The 35 towns are represented by black circles.

We precise by names only the centroids:
$K = 2$: *Berlin, Marseille*
$K = 3$: *Berlin, Luxembourg, Marseille*
$K = 4$: *Barcelona, Kobenhaven, Luxembourg, Wien*
$K = 5$: *Kobenhaven, Luxembourg, Munich, Madrid, Naples*.

In all these examples the clusters seem quite reasonable and can be well delimited by simple ellipses even in non-euclidean cartographic map.

3.2.1.2 Dissimilarity data

The first name (in bold letters) represents the centroid: $K = 2$:

***Saudi Arabia****, France, Germany, Spain, USA, Brazil, Egypt, Syria, Finland, India, China, Thailand, Australia, South Korea*
***Portugal****, Italy, Mexico, Cuba, Argentina, UAE, Turkey, Russia, Sweden, United Kingdom, Japan*

*K = 3*:
***Germany****, France, Spain, USA, Brazil, Egypt, Saudi Arabia, Syria, Australia, South Korea*
***Portugal****, Italy, Mexico, Cuba, Argentina, Russia, Japan*
***Finland****, UAE, Turkey, Sweden, United Kingdom, India, China, Thailand*

*K = 4*:
***Portugal****, Mexico, Argentina, Turkey, Russia, Sweden*
***Australia****, Germany, France, Spain, Brazil, Egypt, China*
***UAE****, Italy, Cuba, United Kingdom, Japan*
***Syria****, USA, Saudi Arabia, Finland, India, Thailand, South Korea*

*K = 5*:
***Australia****, France, Germany, Spain, Brazil, Egypt*
***Saudi Arabia****, USA, Turkey, Syria, Finland, South Korea*
***China****, UAE, India, Thailand*
***Cuba****, Italy, United Kingdom, Japan*
***Portugal****, Mexico, Argentina, Russia, Sweden.*

It seems evident that for *K = 2* the cluster *Portugal* contains elongated countries. However, *Thailand* is not in this cluster: the surface of its elongated south promontory is relatively small and thus *Thailand* is classified by the shape of its main north territory. Analogically Sardinia is too small to classify *Italy* to the cluster of non-elongated countries.

For *K = 3* the cluster *Portugal* of elongated countries is more restricted, probably to "thinner" countries. But it is hard to interpret the difference between the clusters *Germany* and *Finland*. The difference between the clusters *Cuba* and *Portugal* for *K = 5* might be once more interpreted as a difference between the "thin" and the "thick" elongated countries. However, the interpretation of the other three clusters *Australia*, *Saudi Arabia* and *China* is not clear.

### 3.2.2 Multidimensional mapping

#### 3.2.2.1 Road distances data

To map the 35 European towns, we have computed 2D vectors. The quality of this representation is $\|d\|/\|D\| = 99.8\%$ and $\|D-d\|/\|D\| = 5.6\%$. We note that the fact that the sum of the two percentages is not 100 % is quite normal: it is the squared norms which adds to 100 % (Pythagoras theorem). The quality of the 2D approximation is in the figure 3.

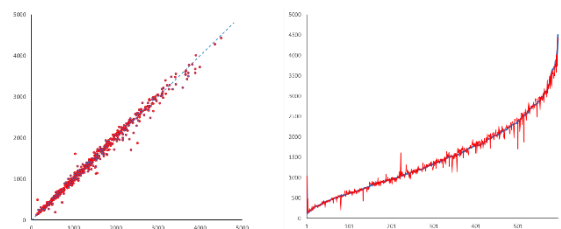

Fig. 3: Two ways how to compare the data matrix (road distances) *D* with the calculated distances matrix d. Left image: the coordinates of the red points are $(D_{i,j}; d_{i,j})$, i, j = 1...N, j > i. The blue dashed line represents the identity $(D_{i,j}; D_{i,j})$. Right image: horizontal axis: rank of the matrix entry; blue line: sorted original data *D*, red line: the corresponding calculated distances *d*. The two plots yield the same information; however, we prefer slightly the right one, which yields some idea about the structure of the data *D*.

The deviations are due to the fact that the road distances are not pure 2D euclidean distances. The corresponding map is in the figure 4.

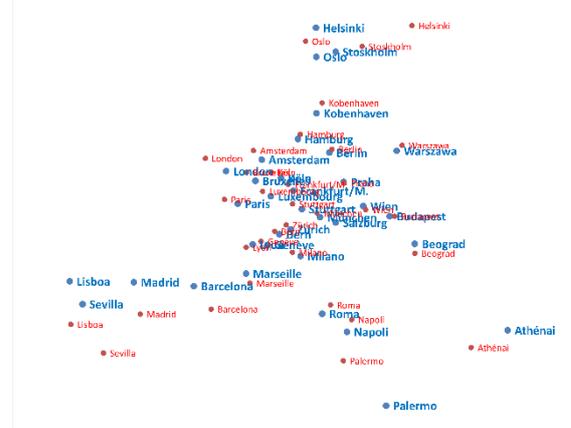

Fig. 4: Euclidean 2D distances map fitted to the real geographical map. Red points: calculated cartesian positions. Blue points: town positions in the geographical atlas.

The calculated coordinates were scaled (by least squares method) to fit the town coordinates of a geographical atlas. The visual agreement could hardly be better, mainly because the geographical map is anything but an euclidean plane.

The result obtained with 3D vectors is in the figure 5.

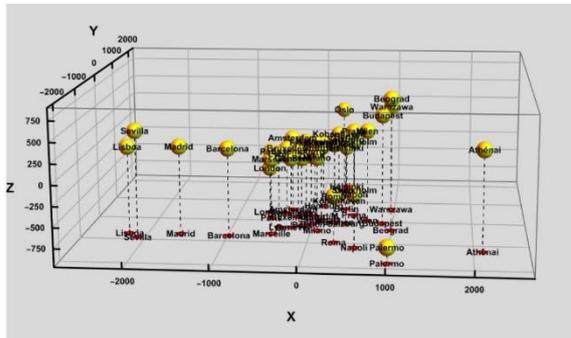

Figure 5: 3D map obtained from the road distances.

The axes were rotated to correspond to principal axes. As expected, the quality of the 3D mapping is better: $\|d\|/\|D\| = 99.9\%$ and $\|D-d\|/\|D\| = 4.7\%$. The spreading in the *z* direction is clearly smaller than that in the *x* and *y* directions. It means that the data correspond approximately to a 2D problem.

Increasing further the dimensionality *n* brings about only minor amelioration: for *n = 5*: $\|d\|/\|D\| = 99.9\%$ and $\|D-d\|/\|D\|= 4.5\%$ and the same values for *n = 10*. The question is if, to make a 2D map, it is better to extract from the data directly 2D vectors or to use higher dimensionality vectors and to project them

(by principal components analysis) to a plane. The answer is unequivocal: the projection of the 3D vectors to the optimal plane yield $\|d\|/\|D\| = 96.4\%$ and $\|D-d\|/\|D\| = 6.9\%$, the projection of the 10D vectors to the optimal plane yields $\|d\|/\|D\| = 85.8\%$ and $\|D-d\|/\|D\| = 16.4\%$, both clearly much worse than using directly the 2D vectors.

3.2.2.2 Dissimilarity data

Increasing the dimensionality $n$ of the extracted vectors ameliorates continuously the quality:

| n | $\frac{\|d\|}{\|D\|}$ [%] | $\frac{\|D-d\|}{\|D\|}$ [%] |
|---|---|---|
| 2 | 97.7 | 21.2 |
| 3 | 99.0 | 14.3 |
| 5 | 99.7 | 7.7 |
| 20 | 100.0 | 0.8 |

Even when the dissimilarity scores don't violate the triangle inequality, neither 2D nor 3D mapping are not good due to the fact that the space disposition of the shapes is far from being bi- or tri-dimensional. The corresponding distances (Fig.6, left image) show clearly visible systematic error. The great deviations concern mostly *Italy*, *Japan*, *Cuba* and *India*. The best plane projections are once again sensibly worse than those obtained directly by extracting directly 2D vectors: the 2D vectors corresponding to the floor of the 3D box yields $\frac{\|d\|}{\|D\|} = 79.0\%$ and $\frac{\|D-d\|}{\|D\|} = 33.7\%$ and the best 2D projection of the 20D vectors yields $\frac{\|d\|}{\|D\|} = 79.6\%$, $\frac{\|D-d\|}{\|D\|} = 30.6\%$. The qualities of the 2D and 20D approximations are compared in the Fig. 6.

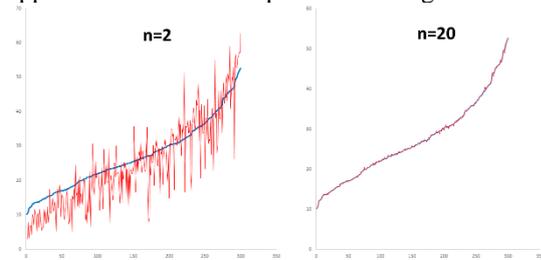

Fig. 6: Quality of multidimensional mapping of the state shapes dissimilarity data for the dimensionality n = 2 and n = 20. Blue line: ordered dissimilarity data **D**, red line: the corresponding calculated distances **d**.

There is no "real" map (as that of Europe for the road distances) to which the results could be compared. Even the 3D mapping (Fig. 7) doesn't explain well the space relations between the shapes. The best plane map which we can have is that obtained with 2D vectors. However, as explained above, *Italy*, *Japan*, *India* and some others are probably very misplaced. $K = 2$ clusters of the block matrix method can still be visualized by simple ellipses, but for $K \geq 3$ some clusters take a "boomerang" or another non-trivial form (see paragraph 3.2.4.2 below). It is due to the fact that the 2D representation is unable to show well the space disposition of the objects.

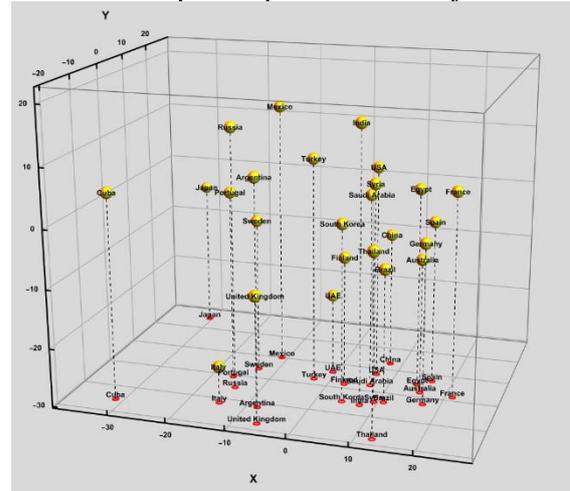

Fig. 7: 3D map of the states obtained from dissimilarity matrix.

### 3.2.3 Torgerson matrix method

3.2.3.1 Road distances data

With the road distance matrix, there are only 19 terms (of 35) which can be used to calculate the cartesian coordinates. If only the first two (greatest) terms of these 19 are used to make a 2D map, we obtain $\frac{\|D-d\|}{\|D\|} = 7.4\%$. It seems natural that increasing the number of terms (and thus the dimensionality n of the vectors) would ameliorate the approximation, but the effect is rather surprising.

| n | $\frac{\|d\|}{\|D\|}$ [%] | $\frac{\|D-d\|}{\|D\|}$ [%] |
|---|---|---|
| 3 | 103.2 | 7.4 |
| 5 | 105.8 | 9.1 |
| 10 | 107.6 | 10.7 |
| 19 | 108.2 | 11.5 |

3.2.3.2 Dissimilarity data

With the dissimilarity data matrix, 21 terms of 25 can be used to calculate the cartesian coordinates. Increasing the dimensionality of the calculated vectors yields:

| n | $\frac{\|d\|}{\|D\|}$ [%] | $\frac{\|D-d\|}{\|D\|}$ [%] |
|---|---|---|
| 2 | 73.1 | 39.6 |
| 3 | 79.9 | 30.4 |
| 5 | 88.1 | 18.6 |
| 10 | 97.3 | 5.8 |
| 21 | 101.1 | 1.7 |

For both the road distances as well as for the dissimilarity data matrices, the results obtained with the Torgerson method are worse that those obtained for the same dimensionality n by multidimensional mapping. Moreover, the deviations are clearly systematic. This is due to the fact that the Torgerson approach solves a transformed problem. The only interest of that method thus seems that it doesn't need any minimization.

### 3.2.4 K-means

3.2.4.1 Road distances data

As with the block-matrix method, K-means yields quite intuitive results, i.e., they can be represented by simple ellipses on the 2D map -see Fig. 8.

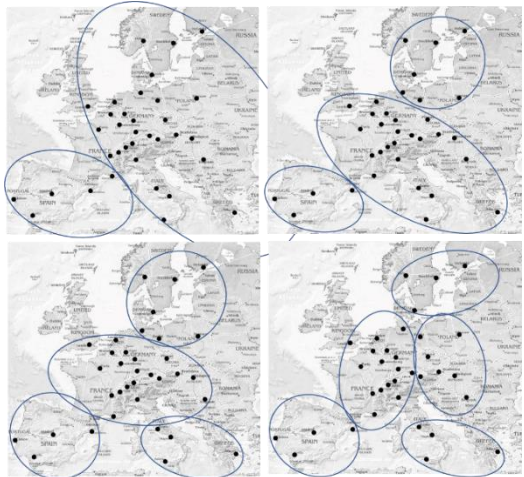

Figure 8: K-means clustering of the road distances matrix for the number of clusters K = 2. . .5 . The towns are represented by black circles.

One can ask why the clusters obtained with the two methods are not the same. The reason is that the methods don't have the same objectives. In the K-means each object joins individually the nearest centroid, in the block matrix method the objects groups together to "cliques". If we name "cohesion" the total sum of all intra-block distances, it is always smaller for the block-matrix method than for the K-means. However, it is hard to tell which of the two methods is "better", and in which sense.

3.2.4.2 Dissimilarity data

The results of K-means method using 20D vectors, compared to the block matrix method, are in the Figures 9 and 10.
The positions of the points correspond to the floor of the 3D box of the Fig. 6. As in the block matrix clustering, the cluster "Portugal" in K-means groups mainly elongated countries
It is interesting to note that for K ≥ 3 the cluster "Japan" in K-means contains only one element. It seems that the cluster "Saudi Arabia" groups "non-elongated" countries, but the overall interpretation of clusters is not evident.

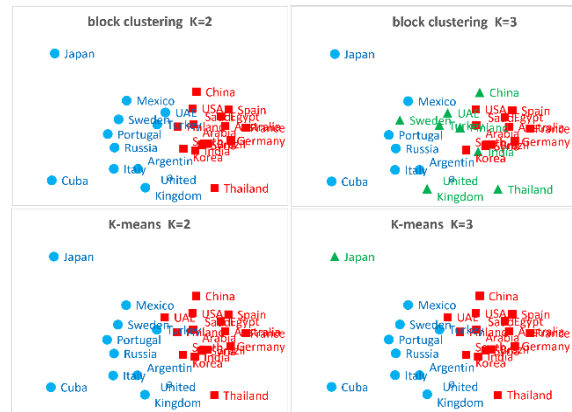

Fig. 9: Comparison of block matrix clustering and K-means

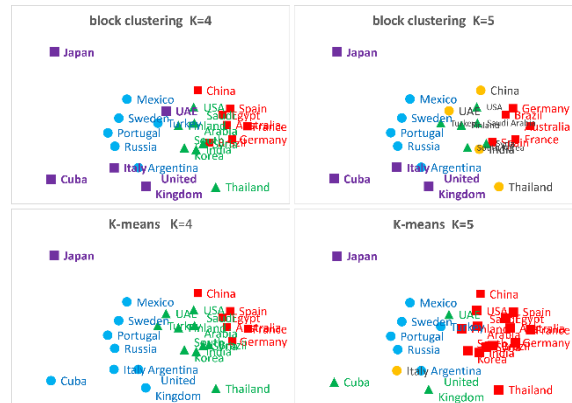

Fig. 10: Comparison of block matrix clustering and K-means

When the clusters yielded by block matrix clustering can be well represented by simple ellipses (see Fig. 2), it is not the case of the dissimilarity scores, where the 2D plane is a very bad representation of the real spatial layout of the shapes. There are regions more complicated than the simple ellipses. For instance, Italy is very badly represented in 2D plane. It is confirmed be the corresponding entries of the difference matrix ($D−d$).

### 3.2.5 Correlation method

3.2.5.1 Road distances data

For the road distance matrix, using 2D vectors, the method yields: r = 0.994, $\frac{\|D-d\|}{\|D\|}$=11.0 % and for 5D vectors yield r = 0.997, $\frac{\|D-d\|}{\|D\|}$=7.5 %. It is worse than with multidimensional mapping, but for n ≥ 3, it is better than the results with the Torgerson matrix with the same dimensionality.

The 2D map is visually identical to that of the 2D mapping in the Fig. 4. However, as was told above, this approach is really interesting only in "desperate" cases, when the data cannot be easily assimilated to the distances.

3.2.5.2 Dissimilarity data.

With 2D vectors the method yields: r = 0.958, $\frac{\|D-d\|}{\|D\|}$=32.7 %, with 5D vectors r = 0.993 and

$\frac{\|D-d\|}{\|D\|}$=11.9 % and with 8D vectors r = 0.997 and $\frac{\|D-d\|}{\|D\|}$=7.4 %. These values are worse than those of multidimensional mapping, but better that those obtained by the Torgerson matrix with the same dimensionality. The 2D maps obtained for n = 2 by multi-dimensional mapping and the correlation method are very different. However, it is not surprising as it has no much sense to compare two very bad approximations.

## 4. CONCLUSION

We have presented an original method to calculate pair dissimilarities among the shapes of a set of 2D figures. The complexity of the shapes (convex/concave, discontinuous, with holes inside) brings no problem. We have used four parameters (size, 2D translation and rotation) to fit the shapes. Many other operations would be possible (flipping over, various deformations) provided that they could be well parameterized. The drawback of the method is that it is rather slow (the program must search and find all the non-overlapping polygons). And to obtain "manually" the points on the contours
with a photo editor is very, very long work. To automatize this would be a big progress.

Among the methods of analysis of the data (distances, dissimilarity) matrix we have presented also two original approaches (block matrix clustering and correlation method). The simple "block matrix clustering" works quite satisfactorily. But it might be much harder (or at least much longer) for much larger matrices.

The multidimensional mapping is very simple and fast minimization problem. To obtain the 2D map, it is much better to search directly 2D vectors than to project multi-dimensional vectors to the plane. However, for the problems like the shapes similarity, the 2D map is a very bad representation of the spatial disposition of the shapes.

The only advantage of using the Torgerson matrix is that it avoids any minimization. However, it is at the price of solving a non-linearly transformed problem. Finally, the "correlation" approach is similar to the multidimensional mapping. It is interesting in the situations where the original data cannot be simply assimilated to distances.

The interpretation of the road distances data by all these methods was simple. It is much less clear with the dissimilarity scores. However, it is not because of these methods, but due to the complexity of the data used (real state shapes).

## Acknowledgements

This paper wouldn't exist without the help of several persons. My wife Tatiana spent several winter evenings by dictating to me the (x,y) coordinates of the points on the figure contours. The nice 3D yellow sphere spreads were made by Guillaume Santini. And the very useful labels (town or state names) on the Excel Graphs were programmed by Michal Reiter. The help of J. F. Gibrat with LateX was essential. And, the last but not least, J. Obdržálek and J. Pothier for their comments to the manuscript.